\documentclass[10pt,twocolumn,letterpaper]{article}

\usepackage{cvpr}

\makeatletter
\@namedef{ver@everyshi.sty}{}
\makeatother
\usepackage{tikz}
\usepackage[utf8]{inputenc} 
\usepackage{times}
\usepackage{epsfig}
\usepackage{graphicx}
\usepackage{amsmath}
\usepackage{amssymb}
\usepackage{appendix}

\usepackage[ruled,vlined]{algorithm2e}
\usepackage{todonotes}
\usepackage{tummath}
\usepackage{tumcolors}  
\usepackage{pgfplots}
\usepackage{subcaption}
\usepackage{booktabs}
\usepackage[inline]{enumitem}
\usepackage[font=small]{caption}

\usepackage[detect-all]{siunitx}
\DeclareSIUnit[product-units=single]\pixel{px}

\usepackage{authblk}

\usetikzlibrary{calc, fit, positioning}
\usetikzlibrary{external}

\colorlet{mamlcolor}{tumorange}
\colorlet{pretrainedcolor}{tumblue}
\colorlet{randomcolor}{tumbluelight}

\newlength{\commentWidth}
\def\reducecaptionspace{}

\usepackage[pagebackref=true,breaklinks=true,letterpaper=true,colorlinks,bookmarks=false]{hyperref}

\usepackage[eulergreek]{sansmath}
\pgfplotsset{
	y tick label style={/pgf/number format/.cd,scaled y ticks = false,set thousands separator={},fixed},
	tick label style = {font=\sansmath\sffamily\footnotesize},
	every axis label = {font=\sansmath\sffamily\scriptsize},
	every axis/.append style={axis lines=left, thick},
	legend style = {font=\sansmath\sffamily\scriptsize, draw=white, fill opacity=.5, text opacity=1},
	label style = {font=\sansmath\sffamily\footnotesize},
	grid style={line width=.1pt, draw=gray!10},
	major grid style={line width=.2pt,draw=tumgraylight},
	title style={font=\sansmath\sffamily},
	grid=both,
}

\usepackage[capitalize]{cleveref}

\cvprfinalcopy 

\def\cvprPaperID{45} 

\ifcvprfinal\fi

\newcommand\blfootnote[1]{%
  \begingroup
  \renewcommand\thefootnote{}\footnote{#1}%
  \addtocounter{footnote}{-1}%
  \endgroup
}

\begin{document}
	
	\title{Meta-Learning for Few-Shot Land Cover Classification}
	%
	%
	
    \author[1,*,$\dagger$]{Marc Rußwurm}
    \author[2,3,*]{Sherrie Wang}
    \affil{Technical University of Munich, Chair of Remote Sensing Technology}
    \author[1]{Marco Körner}
    \affil{Stanford University, Center on Food Security and the Environment}
    \author[2]{David Lobell}
    \affil{Stanford University, Institute for Computational and Mathematical Engineering}

	\maketitle
	
	\begin{abstract}
		The representations of the Earth's surface vary from one geographic region to another.
        For instance, the appearance of urban areas differs between continents, and seasonality influences the appearance of vegetation. 
        To capture the diversity within a single category, like as urban or vegetation, requires a large model capacity and, consequently, large datasets.
        In this work, we propose a different perspective and view this diversity as an inductive transfer learning problem where few data samples from one region allow a model to adapt to an unseen region.
        We evaluate the model-agnostic meta-learning (MAML) algorithm on classification and segmentation tasks using globally and regionally distributed datasets.
        We find that few-shot model adaptation outperforms pre-training with regular gradient descent and fine-tuning on (1) the Sen12MS dataset and (2) DeepGlobe data when the source domain and target domain differ.
        This indicates that model optimization with meta-learning may benefit tasks in the Earth sciences whose data show a high degree of diversity from region to region, while traditional gradient-based supervised learning remains suitable in the absence of a feature or label shift. 
	\end{abstract}
	
    \blfootnote{*equal contribution; $\dagger$ this work was conducted during a research stay at the Lobell Lab and supported by a fellowship within the IFI programme of the German Academic Exchange Service (DAAD). }

	\section{Introduction}
	\label{sec:intro}

	\begin{figure}[t]
%
		\ifdefined\reducecaptionspace
		\setlength{\belowcaptionskip}{-15pt}
		\fi
		
%
			\includegraphics{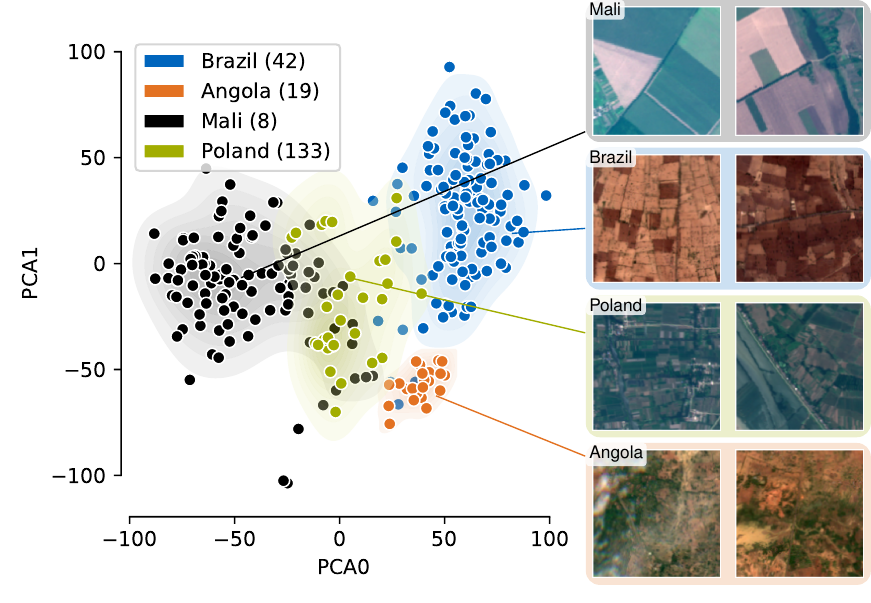}
			\caption{A principle components analysis (PCA) on VGG-16~\cite{simonyan2014very} features of cropland images from different countries. Representations of the same class vary geographically; applying models trained on one geography to another would violate the assumption in traditional supervised learning that train and test distributions are equal. Model-agnostic meta learning provides a framework for inductive transfer learning that adapts the model to a new region with few data samples.}
			\label{fig:representations}

%
	\end{figure}

	A growing constellation of satellites, combined with cloud computing and deep learning, offers an objective and scalable way to monitor global issues from deforestation and wildfires to urban development and road flooding~\cite{mccracken1999remote,chuvieco2012remote,bhatta2010analysis,schnebele2014road}. 
    For many of these prediction problems, the bottleneck to making accurate and timely predictions has shifted away from satellite imagery availability or data processing limits and toward a lack of ground truth labels~\cite{zhu2017deep,wang2020weakly,scott2017training}. 
    At the same time, these tasks share characteristics in remotely sensed imagery---such as ground sampling distance, seasonality, and spectral characteristics---no matter where on Earth they are taken. 
    This raises the question of whether prediction in label-scarce regions could be improved if each model were to benefit from knowledge contained in all the datasets, rather than solving the same prediction problem across different geographies or time slices with independent models trained on small disjoint datasets.
    
    The concept of using knowledge gained while solving one problem to aid the solving of another is known in machine learning as \textbf{transfer learning}~\cite{pan2009survey}.
    Transferring knowledge between tasks or domains is successful when the problems are different but related~\cite{tuia2016domain}.
    We argue that the diverse nature of representations on the Earth's surface is a prime example of different-but-related tasks.
    We illustrate this in \cref{fig:representations} using representations of cropland from four different countries.
    Croplands across the world are distinct from each other, yet they share characteristics. 
    Transfer learning allows models to both adapt to each distribution individually and share knowledge across regions: countries like Angola and Mali, for which smaller labeled datasets are available, could then benefit from larger labeled datasets from countries like Brazil and Poland.
    
\begin{figure}
		
	\ifdefined\reducecaptionspace
	\setlength{\belowcaptionskip}{-15pt}
	\fi

	\includegraphics{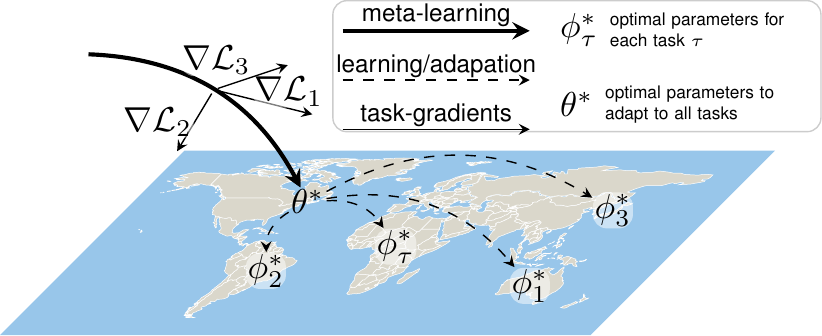}
	\caption{The model-agnostic meta learning (MAML) algorithm \cite{maml} finds initial weights $\theta$ from which a model can adapt to a new geographic region $\tau$ with few data samples.}
	\label{fig:metalearning}
\end{figure}

    Thus far, transfer learning on remote sensing data has largely focused on fine-tuning pre-trained models and performing domain adaptation (Section \ref{sec:related}). 
    In this work, we explore \textbf{meta-learning}, in which models not only learn from data to perform tasks but \emph{learn how to learn} to perform tasks through experiencing tasks on a variety of datasets. 
    In particular, we use model-agnostic meta-learning (MAML) for the problem of inductive transfer-learning, where the generalization is induced by a few labeled examples in the target domain~\cite{pan2009survey}.
    A schematic of MAML is shown in \cref{fig:metalearning} and the algorithm is described in \cref{sec:maml}.
    
    Our main contributions are (1) demonstrating that remote sensing tasks across geographies can be restructured as one meta-learning problem and 
    (2) evaluating MAML for few-shot classification and segmentation of multi-spectral and high-resolution remote sensing images; specifically, the well-cited benchmark datasets Sen12MS and DeepGlobe.
	
	
%
	\section{Related Work}
	\label{sec:related}
	
	Transfer learning can be divided into subcategories depending on the amount of labeled data available in the source and target domains. 
	Our work is focused on the scenario in which ample labels exist in the source domain, but few exist in the target domain. 
	We summarize the related remote sensing methodology accordingly. 
	
    
    
    In such a setting, one common transfer learning technique is pre-training a neural network on ImageNet and fine-tuning~\cite{marmanis2015deep} it on an application-specific dataset.
    For high-resolution remotely sensed imagery, these include airplane detection \cite{chen2018end}, high-resolution land cover classification \cite{tong2020land}, and disaster mapping \cite{gupta2019post}.
    Xie et al. (2016)~\cite{xie} extended this concept by swapping ImageNet for the proxy task of night-light prediction that allowed them to estimate poverty in African regions with a limited number of labeled poverty data points.
    These approaches require a significant amount of problem design, such as the choice of proxy datasets or model and which parameters to fine-tune, and, thus, usually focus on a limited number of hand-selected tasks. 
    
    A second class of methods using deep learning for label-scarce tasks in remote sensing has focused on developing novel network architectures or loss functions to make learning more label-efficient. 
    So far, these methods have focused on optical~\cite{jean2019tile2vec}, SAR~\cite{rostami2019deep}, and hyperspectral image classification~\cite{liu2019deep}. 
    While they decrease the number of labels required for any optical, SAR, or hyperspectral task, these methods do not explicitly endeavor to transfer knowledge from a data-rich geography to a data-poor one.
    
    Non-deep learning methods for domain adaptation were summarized by Tuia et al. (2016)~\cite{tuia2016domain} and include selecting invariant features, adapting data distributions, and adapting classifiers via semi-supervised learning. 
    For the most part, such methods generalize only across small regions rather than worldwide, while sometimes requiring a feature space in which inputs can be modeled as a mixture of Gaussians or some other predefined distribution.
    
    Lastly, meta-learning is beginning to be explored for remote sensing applications.
    Alajaji and Alhichri (2020)~\cite{alajaji2020few} describe preliminary results of MAML on few-shot UC Merced, OPTIMAL-31, and AID RS classification, though again not with a focus on cross-geography generalization.
	
	
	\section{Meta-learning}
	\label{sec:metalearning}
	
	Meta-learning \cite{schmidhuber1987evolutionary} considers a large number of related tasks $\tau \in \mathcal{T}=\{\tau^{(1)},\dots,\tau^{(N)}\}$ 
	to arrive at a predictive function that can perform well on 
	unseen tasks $\tau$ after seeing a few data samples. 
    Even though meta-learning has been a topic in machine learning for decades  \cite{schmidhuber1987evolutionary,bengio1990learning}, it has recently gained popularity for few-shot problems \cite{vinyals2016matching,snell2017prototypical,ravi2016optimization} and has been re-introduced under a ``model agnostic'' framework~\cite{maml} with rapid developments in the field~\cite{rajeswaran2019meta,nichol2018first,antoniou2018train}.
	
	\begin{algorithm}[t]
	\SetAlgoLined
	$p(\mathcal{D})$: distribution over data points\;
	$\alpha$: step size hyperparameters\;
	randomly initialize $\phi$\;
	\Repeat{convergence}{
		sample ${D} \sim p(\mathcal{D})$\;
		evaluate $\V{g}=\nabla\mathcal{L}(f_{\phi},{D})$\;
		update parameters $\phi \leftarrow \phi - \alpha\V{g}$\;
	}
	\caption{Regular Gradient Descent}
	\label{alg:grad_descent}
\end{algorithm}
    \begin{algorithm}[t]
	\SetAlgoLined
	$p(\mathcal{T})$: distribution over tasks\; $\alpha, \beta$: step size hyperparameters\;
	randomly initialize $\theta$\;
	\Repeat{convergence}{
		sample batch of tasks $\tau \sim p(\mathcal{T})$\;
		\ForEach{$\tau_{i} \in \tau$}{
			initialize $\phi_{i}$ with $\theta$\;
			sample $\{{D}_\text{support}, {D}_\text{query}\} \sim p(\tau_i)$\;
			evaluate $\V{g}=\nabla_{\phi_{i}}\mathcal{L}_{\tau_{i}}(f_{\phi_{i}},{D}_\text{support})$\;
			adapt parameters $\phi_{i} \leftarrow \phi_{i} - \alpha\V{g}$\;
			evaluate test loss $\mathcal{L}_{\tau_{i}}(f_{\phi_{i}},{D}_\text{query})$ \;
		}{
			update $\theta \leftarrow \theta - \beta \sum_{\tau_{i} \sim p(\tau)} \nabla_{\theta}\mathcal{L}_{\tau_{i}}(f_{\phi_{i}},{D}_\text{query}^{\tau_i})$\;
		}
	}
	\caption{Model-Agnostic Meta-Learning}
	\label{alg:maml}
\end{algorithm}
    
	\subsection{Terminology and Definitions}
	Meta-learning 
	introduces a set of terms that may be new to some readers, so we clarify them in this section.
	
		A \textbf{task} $\tau$ is comprised of a \textbf{support} dataset ${D}_\text{support}$ to adjust the model parameters to the specific task and a \textbf{query} dataset ${D}_\text{query}$ to evaluate the performance. 
		Each dataset is comprised of inputs $\{\V{x}_{1}, \V{x}_{2}, \hdots, \V{x}_{m}\}$ and corresponding labels $\{y_{1}, y_{2}, \hdots, y_{m} \}$ from a data distribution. 
		A \textbf{$k$-shot}, \textbf{$n$-way} classification task aims to distinguish between $n$ classes and is trained on $k$ examples per class. 
		Each task is drawn from a distribution over tasks $\tau \sim p(\mathcal{T})$ to yield a set of tasks $\{\tau^{(1)},\tau^{(2)},\dots, \tau^{(N)}\}$. 
		The meta-learner \emph{learns how to learn} by training and evaluating on the \textbf{meta-training set}.
		Meta-learning hyperparameters are tuned on the \textbf{meta-validation set}. 
		The \textbf{meta-test set} measures generalization on new, unseen tasks. 

	\subsection{Model-Agnostic Meta Learning (MAML)}
	\label{sec:maml}
	Neural network parameters $\phi$ are usually initialized randomly and optimized iteratively via gradient descent to perform well on a single dataset, as shown in \cref{alg:grad_descent}.
    Model-agnostic meta-learning (MAML) extends gradient descent by optimizing for a model initialization $\theta$ that leads to good performance on a set of related tasks $\{\tau^{(1)}, \tau^{(2)}, \hdots, \tau^{(N)}\}$.
    We contrast the regular gradient descent with the MAML optimization algorithm in \cref{alg:grad_descent,alg:maml}. 
    Meta-training is divided into an inner loop and an outer loop.
    In the inner loop, networks initialized with $\theta$ are updated to each task via $t$ steps of gradient descent on $D_\text{support}$ of each task. 
    This results in models with parameters $\phi_{i}$ adapted to each task $\tau^{(i)}$. 
    The outer loop updates $\theta$ based on the performance of $\phi_{i}$ on $D_\text{query}$ of the meta-training batch.
    In so doing, MAML requires second-order gradient calculations.
    The algorithm looks for a better $\theta$ until convergence, upon which the generalization error is computed on unseen meta-test tasks.

	\section{Datasets}
	
	\begin{figure*}
	\begin{subfigure}[t]{.56\linewidth}
		
	    \includegraphics{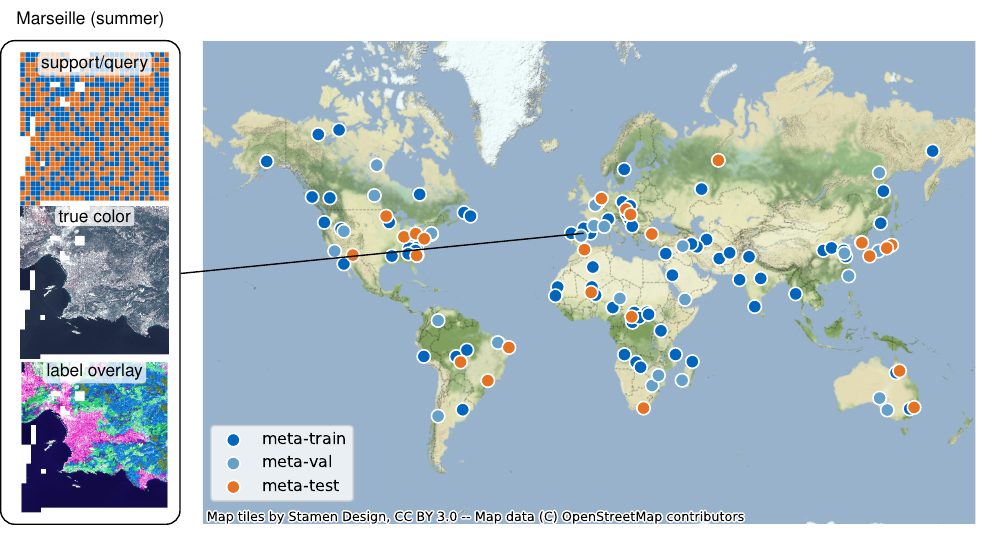}
		
		\caption{The 125 regions of the Sen12MS dataset. The 25 meta-test regions have been selected based on the hold-out set of the Data Fusion Contest 2020~\cite{Yokoya2020}. The 75 meta-train and 25 meta-val have been randomly randomly partitioned.}
		\label{fig:sen12ms:regions}
	\end{subfigure}
	\hfill
	\begin{subfigure}[t]{.42\linewidth}
	    \centering
	    \resizebox{\linewidth}{!}{
	    	\includegraphics{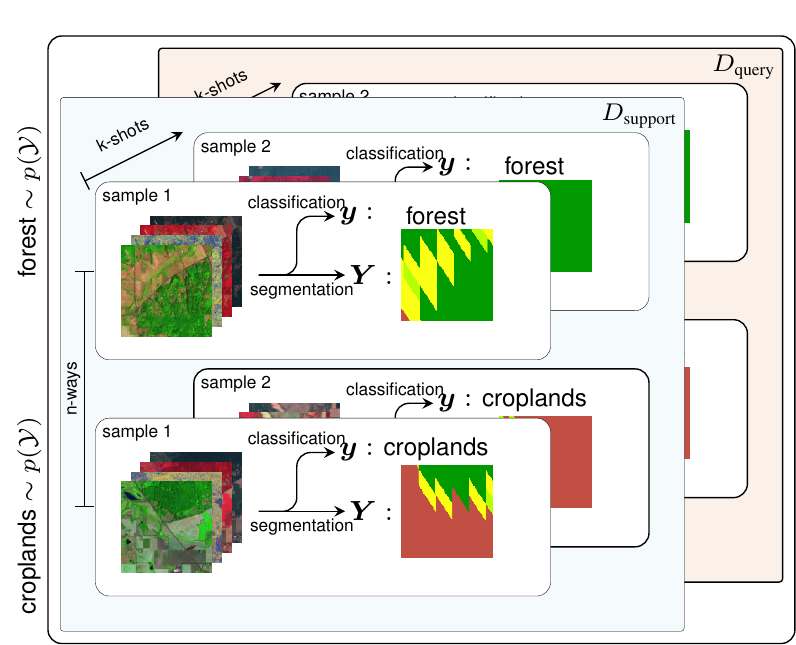}
		}
		\caption{Example of a Sen12MS 2-way-2-shot task from region 87 and in the summer season. Ways determines the number of classes per task while shot the number of samples per class. 
		}
		\label{fig:sen12ms:shotexample}
	\end{subfigure}

	\ifdefined\reducecaptionspace
	\setlength{\belowcaptionskip}{-15pt}
	\fi

	\caption{The Sen12MS dataset~\cite{schmitt2019sen12ms} is a public remote sensing dataset of 128 globally distributed regions and four distinct seasons. In this work, we sample tasks (b) from the dataset that include samples from one region and season aiming at adapting a deep learning model to one specific region.} 
	\label{fig:sen12ms}
\end{figure*}

	\begin{figure}

	\ifdefined\reducecaptionspace
	\setlength{\belowcaptionskip}{-15pt}
	\fi
	
\centering
\includegraphics[width=0.45\textwidth]{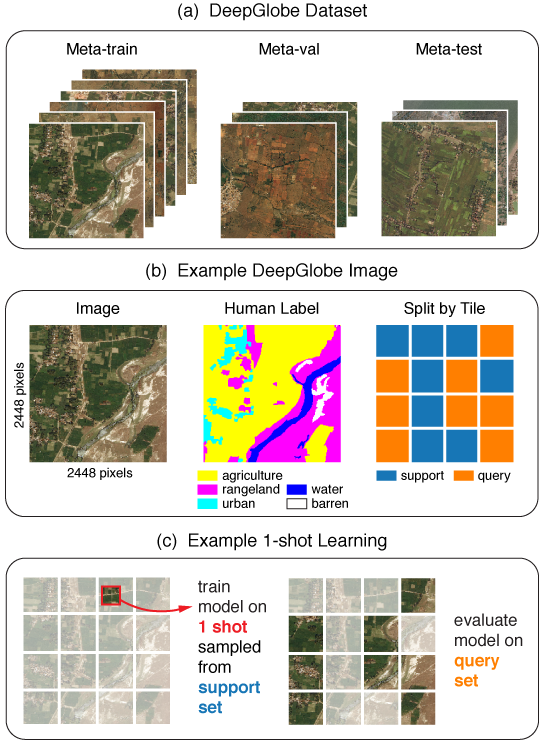}
\caption{The DeepGlobe dataset contains high resolution RGB satellite imagery with land cover labels segmented by humans. To repurpose DeepGlobe for meta-learning, we (a) split the images into meta-train, meta-val, and meta-test sets. Then (b) each image was split into 16 sub-images, 8 of which were placed in the support set and 8 in the query set. Under such a setup, (c) we trained models on the meta-train set to segment the queries after seeing $k$ shots from the support. }\label{fig:deepglobe_dataset}
\end{figure}
	
	We evaluate model-agnostic meta-learning on two public remote sensing datasets that cover optical and radar data at medium and very high resolution.
    
    \subsection{Sentinel-1/2 Multi-Spectral (Sen12MS) Dataset}
    
    The \emph{Sentinel-1/2 Multi-Spectral (Sen12MS)}~\cite{schmitt2019sen12ms} dataset is a novel globally distributed satellite image classification and segmentation dataset.
    It contains \num{280662} Sentinel 2 (optical) and Sentinel 1 (radar) tiles from 125 distinct regions at four different seasons. 
    The optical and radar images were resampled to \SI{10}{\meter} ground sampling distance and span \SI{256 x 256}{\pixel} in height and width.
    The original dataset uses tile-overlaps of 50\%. 
    For this work, we removed the overlap to ensure independence of support and query datasets, which yielded \num{200306} \SI{128 x 128}{\pixel} tiles.
    We show true color examples and principal component embeddings on VGG-16 features of four distinct regions in \cref{fig:representations}.
    Each image tile is accompanied by a land cover label with a comparatively coarse resolution of \SI{500}{\meter} from the MODIS Land Cover product MCD12Q1 V6 upsampled to \SI{10}{\meter}. 
    In this work, we use the Sen12MS dataset for classification and assign the most common pixel-level label to the image tile.
    We use the simplified label-scheme of International Geosphere Biosphere Programme (IGBP) categories~\cite{loveland1997igbp} with 10 distinct classes, consistent with the IEEE Data Fusion Contest 2020~\cite{Yokoya2020}.
    In \cref{fig:sen12ms:regions}, the 125 globally distributed regions are shown separated into meta-train, meta-validation, and meta-test sets. 
    Each region contains between 196 and 850 tiles with a region-specific class distribution. 
    We also show an overview of all tiles of the region 131 (Marseille) from the summer season true-color and labels. 
    The individual \SI{128x128}{\pixel} tiles are randomly assigned to the support or query partition of each region.
    The objective is to classify each tile with its most frequent label class.
    \Cref{fig:sen12ms:shotexample} illustrates this on an example of a 2-shot 2-way task.
    In this case, task-datasets $D_{query}^\tau$ and $D_{support}^\tau$ contain $k=2$ randomly chosen tile-label pairs of $n=2$ distinct classes chosen from the available classes in the region.
	
	\subsection{DeepGlobe Land Cover Segmentation Dataset}
	
	The DeepGlobe Challenge~\cite{demir2018deepglobe} was introduced at CVPR 2018 to advance state-of-the-art satellite image analysis. 
	Here, we used the land cover segmentation data to explore the use of MAML on high-resolution satellite imagery.
    
    The DeepGlobe land cover segmentation dataset is comprised of very high resolution (\SI{0.5}{\meter}) DigitalGlobe Vivid+ images of dimension \SI{2448x2448}{\pixel} with three RGB channels. 
    In total, there are 803 training images, each with human-annotated semantic segmentation labels covering seven land cover classes: urban, agriculture, rangeland, forest, water, barren, and unknown. 
    For the competition, 171 validation images and 172 test images were also provided.
    However, since they do not have corresponding labels, we did not include them in the following experiments.
    Across the training images, the most common class is 
    agriculture (\SI{58}{\percent} of pixels), 
    followed by forest (\SI{11}{\percent}), 
    urban (\SI{11}{\percent}), 
    rangeland (\SI{8}{\percent}), 
    barren (\SI{8}{\percent}), 
    water (\SI{3}{\percent}), 
    and unknown (\SI{0.05}{\percent}).
	
	
	We divided the DeepGlobe training set into three meta-datasets: a meta-training set on which to train MAML, a meta-validation set on which to tune MAML hyperparameters, and a meta-test set on which to evaluate generalization (\cref{fig:deepglobe_dataset}a).
	Ideally, we would evaluate whether meta-learned models generalize better to new geographic regions.
	However, the DeepGlobe Land Cover dataset does not tag images with latitude and longitude.
	In the absence of geographic information, we split the images in two ways:
    \begin{enumerate}
        \item At random, \ie the 803 images were sampled uniformly at random into a 500-image meta-train, a 150-image meta-val, and a 153-image meta-test set.
        \item Using unsupervised clustering on features extracted from a pre-trained network. DeepGlobe images were fed into a VGG-16 network pre-trained on ImageNet, and for each image, a 4096-dimensional vector was extracted from the first layer in the classifier. We used $k$-means to assign the images into 6 clusters and the 6 clusters were divided at random into the meta-train, meta-val, and meta-test sets. The resulting datasets contained 454, 166, and 183 images, respectively.
    \end{enumerate}
    \Cref{fig:deepglobe:pca} visualizes the distributions of image features for the meta-train, meta-val, and meta-test sets under these two splitting methodologies. 
    The results across the two splits will illuminate the settings under which MAML improves upon pre-training and training from scratch.
    
    Each image was further divided into 16 sub-images, each of dimension \SI{612x612}{\pixel} (\cref{fig:deepglobe_dataset}b). 
    Eight sub-images were placed in the support set and 8 in the query set.
    At meta-train time, $k$ shots of \SI{306x306}{\pixel} tiles were sampled from the support set and $q$ queries were sampled from the query set.
    At meta-test time, the entire query set was fed into the model as 32 tiles to compute metrics (\cref{fig:deepglobe_dataset}c). 
    
    Put succinctly, our DeepGlobe experiments explore whether a model can learn to segment a large region (\SI{1.2x1.2}{\kilo\meter}) of high resolution satellite imagery after seeing only a small labeled section (\SI{153x153}{\meter}) of it.

	\section{Models}
	
	Model-agnostic meta-learning is an optimization algorithm that uses gradient descent and can be employed for any neural network architecture.
    In this work, we chose two popular models for image classification and segmentation.
    
    \subsection{CNN Classification Model}
    \label{sec:model:cnn}
    
    Following other meta-learning approaches \cite{maml,vinyals2016matching}, we used a straightforward CNN architecture for the Sen12MS classification objective.
    The network consisted of 7 stacked layers with 64 convolutional \SI{3x3}{\pixel} kernels followed by batch normalization~\cite{ioffe2015batch}, ReLU activation function, and max-pooling of size 2.
    The input tensor $\M{X} \in \R^{128 \times 128 \times 15}$ of joint Sentinel-2 and Sentinel-1 bands is projected to a 64-dimensional feature vector that maps to the output vector $\V{y} \in \R^{10}$ for each of the classes.
    
    \subsection{U-Net Segmentation Model}
    \label{sec:model:unet}
    
    For the DeepGlobe segmentation task, we employed the popular U-Net~\cite{ronneberger2015u} architecture. 
    It is a fully-convolutional segmentation model with skip connections between encoder and decoder. 
    We used four downsampling and upsampling layers so that the input tensor is projected to a hidden representation, which is then added to intermediate hidden states from the encoder (skip connections) while being upsampled and convolved to an output tensor whereupon each pixel represents one class label.
	
	\section{Experiments}
	
	We experimentally evaluated the classification and segmentation performance of deep learning models with the same architecture trained with regular gradient descent (pre-trained) \cref{alg:grad_descent} and MAML \cref{alg:maml}.
    
    \subsection{Sen12MS Classification}
    \label{sec:sen12msclassification}
    
    \tikzsetnextfilename{sen12ms-classification-segmentation}
\begin{figure}
    	\includegraphics{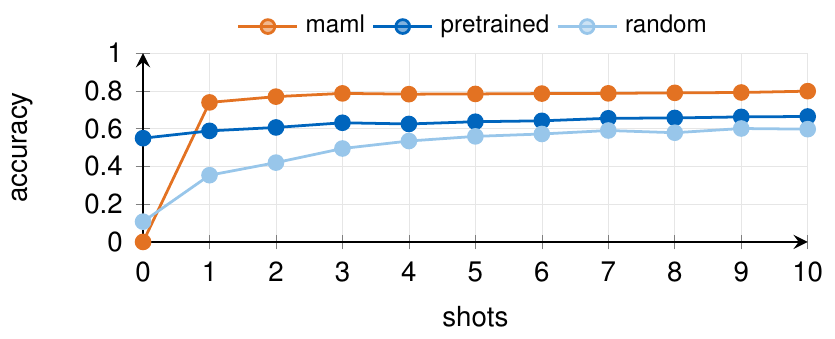}
    	

		\label{fig:exp:classification}
	\ifdefined\reducecaptionspace
	\setlength{\belowcaptionskip}{-15pt}
	\fi
	\caption{Classification results on Sen12MS. Regular pre-training with gradient descent leads to good zero-shot performance, while models trained with the model-agnostic meta learning algorithms outperform regular pretraining and the randomly initialized baseline clearly throughout all ten seen examples from a unseen region.}
	\label{fig:sen12ms-classification-segmentation}	
\end{figure}
    
    We assumed that data from the meta-train regions were readily available, but at most ten image-label pairs per class can be seen from the meta-test regions. 
    This corresponds to a 4-way 10-shot classification scenario with four randomly selected classes from one region. 
    It reflects use-case of interest to this work, where labeled data is available in some regions but not in others.
    
    We trained the classification models with MAML on 4-way 2-shot datasets from the meta-train regions.
    We treated each sub-dataset $D_\text{support}$ and $D_\text{query}$ as a single batch of $N=k \cdot n=8$ samples.
    
    \textbf{Baselines.} We compared the \emph{MAML-trained} model with a model that was {pre-trained} on all available data from the meta-train regions using regular gradient descent \cref{alg:grad_descent}. 
    We \emph{pre-trained} this model with the same 4-way 2-shot batches as MAML but used the combined support and query task-datasets for training. 
    This resulted in a batch size of 16 image label pairs.
    Finally, we also considered the scenario of having no additional data from meta-train regions.
    Here, we initialized the model randomly without any prior training, and train on each task's support set from scratch; we refer to this baseline as the \emph{random} model.
	
	\textbf{Evaluation}. 
    With the three initial CNN model parameterizations, \ie \emph{MAML-trained}, \emph{pretrained}, and \emph{random}, we evaluated the ability to adapt to new unseen meta-test regions based on at most ten data samples.
    For this, we sampled 100 4-way 10-shot task-datasets from the meta-test regions.
    We report performance metrics on $D_\text{query}$ on all ten examples per class but fine-tuned the models on subsets of $D_\text{support}$.
    We also varied the number of seen samples from $D_\text{support}$ incrementally from zero-shot to ten-shot.
    Zero-shot represents no fine-tuning and shows the performance that can be obtained solely based on data from the meta-train regions. 
    Training on batches of one-shot to ten-shot provides incrementally more data from the target region to the models.
    The meta-val regions were used to determine a suitable step size $\alpha \in \{0.001,0.0025,\dots,0.5,0.75,1\}$ and gradient steps on the same data batch $n \in \{1,2,5,10,50,100\}$ for fine-tuning the pre-trained model.
    We evaluated these hyperparameters via grid search for each shot independently. 
        
    \textbf{Classification Results}. 
    In \cref{fig:exp:classification}, we report the accuracy scores for an increasing number of shots.
    The zero-shot case, without any adaptation to the particular meta-test region, shows that the regular pre-trained model performed best with \SI{55}{\percent} accuracy and a kappa score of $0.47$.
    Without any adaptation on the target-region, MAML predictions are low in accuracy, which highlights a distinct difference between meta-learning and pre-training.
    However, when a single data sample from the meta-test region is provided (1-shot), the MAML-trained model (\SI{74}{\percent} accuracy, $0.68$ kappa score) outperforms the pre-trained model (\SI{59}{\percent} accuracy, $0.51$ kappa score) by a large margin.
    The pre-trained model only shows a comparatively slight increase in accuracy (\SI{54}{\percent} to \SI{66}{\percent}) throughout all seen examples while the MAML-trained model scores \SI{80}{\percent} accuracy and $0.76$ kappa score with all 10 shots.

 	
	\subsection{DeepGlobe Land Cover Segmentation}
	\label{sec:deepglobe}
	
	\begin{figure}
    \begin{subfigure}[b]{\linewidth}
		\includegraphics[width=\textwidth]{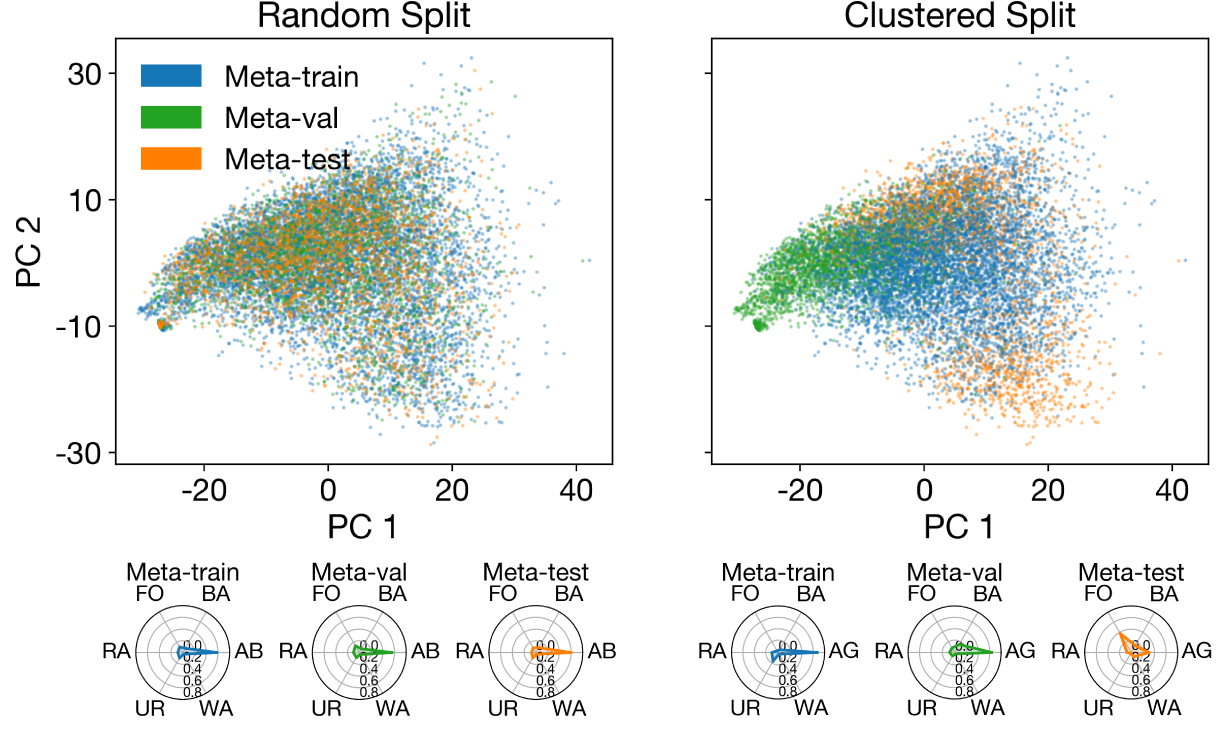}
		\caption{The DeepGlobe images were split into meta-datasets (left) at random, or (right) in clusters based on a lower dimensional representation. Tile representations extracted from a pretrained VGG-16 are plotted along their first 2 principal components. The label distributions of each meta-dataset are shown below.}
		\label{fig:deepglobe:pca}
	\end{subfigure}

	\begin{subfigure}[b]{\linewidth}
		\includegraphics[width=\textwidth]{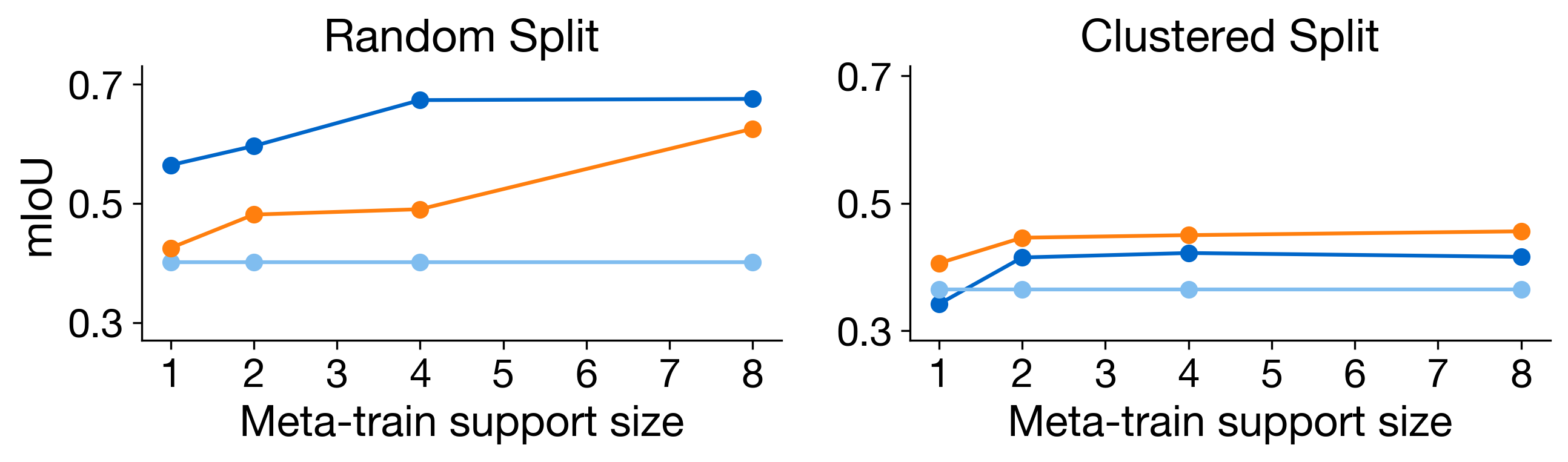}
		\caption{The effect of meta-train support size on segmentation results (mIoU) for (left) randomly split meta-datasets and (right) clustered split meta-datasets. Results are shown for 1 meta-test shot.}
		\label{fig:deepglobe:traintiles}
	\end{subfigure}
	
	\begin{subfigure}[b]{\linewidth}
		\includegraphics[width=\textwidth]{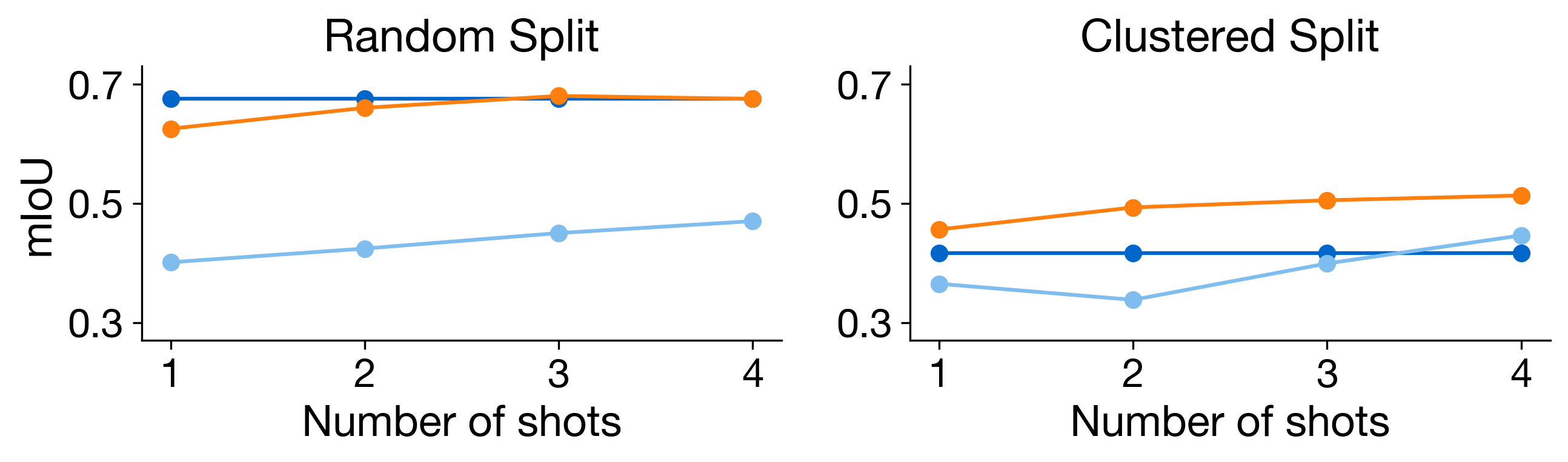}
		\caption{The effect of number of adaptation shots on segmentation results. Results are shown for a support size of 8.}
		\label{fig:deepglobe:shots}
	\end{subfigure}

	\ifdefined\reducecaptionspace
	\setlength{\belowcaptionskip}{-15pt}
	\fi
	\caption{Segmentation results on DeepGlobe, with two ways of splitting the images into meta-datasets.}
	\label{fig:deepglobe_results}	
\end{figure}
	
	Our second experiment demonstrates the use of MAML on the DeepGlobe land cover segmentation dataset. 
    Each DeepGlobe image was considered its own task and we trained a U-Net via MAML to segment the query set of an image after being shown $k$ shots from the support set. 
    The experiments were designed to investigate the effect on the generalization of 
    (1) meta-training label quantity (number of support and query sub-images), 
    (2) meta-test label quantity (number of shots),
    and (3) distributional shift between meta-train and meta-test sets (random split versus clustered split of meta-datasets). 
    The number of labeled sub-images in the support and query sets was varied to be $m \in \left\{1, 2, 4,8\right\}$ and the number of shots used to adapt the U-Net was in the range $k\in \left\{1, 2, 3, 4,5\right\}$. 
    Hyperparameters, such as the number of epochs to meta-train MAML or train a model from scratch, were selected using performance on the meta-validation set.
    
    \textbf{Baselines}.
    Similar to the Sen12MS evaluation, we compared MAML to two baselines: 
    (1) a U-Net pre-trained on the meta-training set and fine-tuned on $k$ shots in each meta-test task, and 
    (2) randomly initialized U-Nets trained independently from scratch on $k$ shots in each meta-test task. 
    To make comparisons fair, we showed the pre-trained model the same amount of data as seen by MAML. 
    If MAML was meta-trained on $m$ support tiles and $m$ query tiles and adapted using $k$ shots, the baseline U-Net was pre-trained on $2m$ tiles per image and fine-tuned on $k$ shots per meta-test tile, and the randomly initialized model was trained on $k$ shots. 
    The U-Net architecture was shared among MAML and both baselines.
    
    \textbf{Evaluation}. 
    The performance of all models was evaluated on the query tiles of an unseen meta-test set of images. 
    The location of the $k$ shots for each meta-test image was sampled at random from its support set and fixed across all models for direct comparison. 
    The models were evaluated by means of pixel-wise accuracy and the mean intersection over union (mIoU) score across the meta-test queries.
    For elaboration on the formula used to compute mIoU, please refer to the DeepGlobe publication \cite{demir2018deepglobe}.
	
	\begin{figure}
    
	\begin{subfigure}[b]{\linewidth}
		\includegraphics[width=\textwidth]{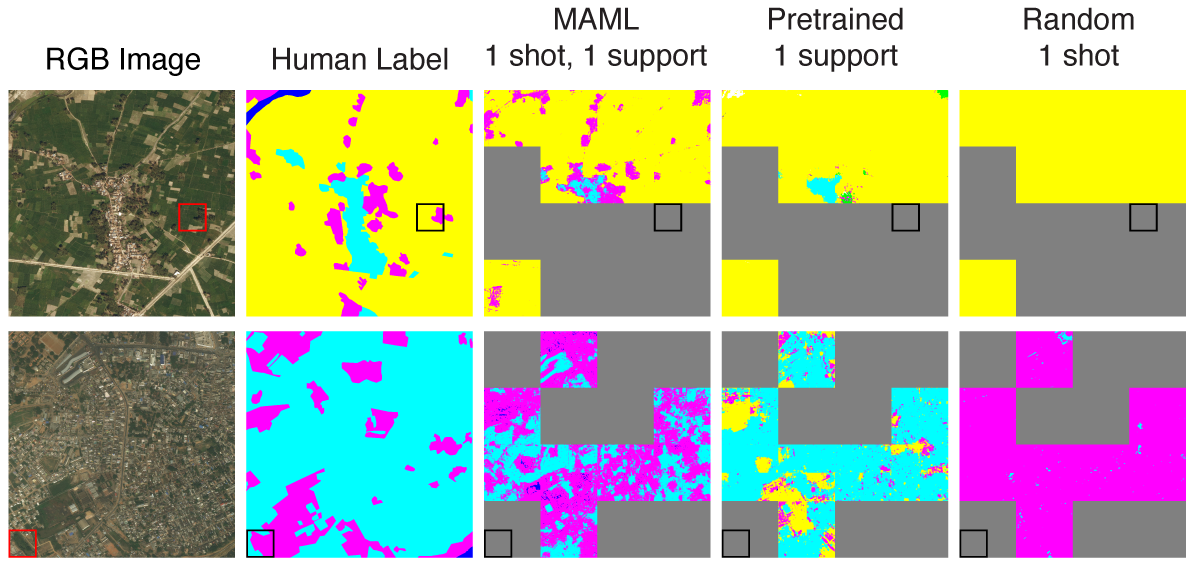}
		\caption{Random meta-dataset splits}
		\label{fig:deepglobe:examples_random}
	\end{subfigure}
	
	\begin{subfigure}[b]{\linewidth}
		\includegraphics[width=\textwidth]{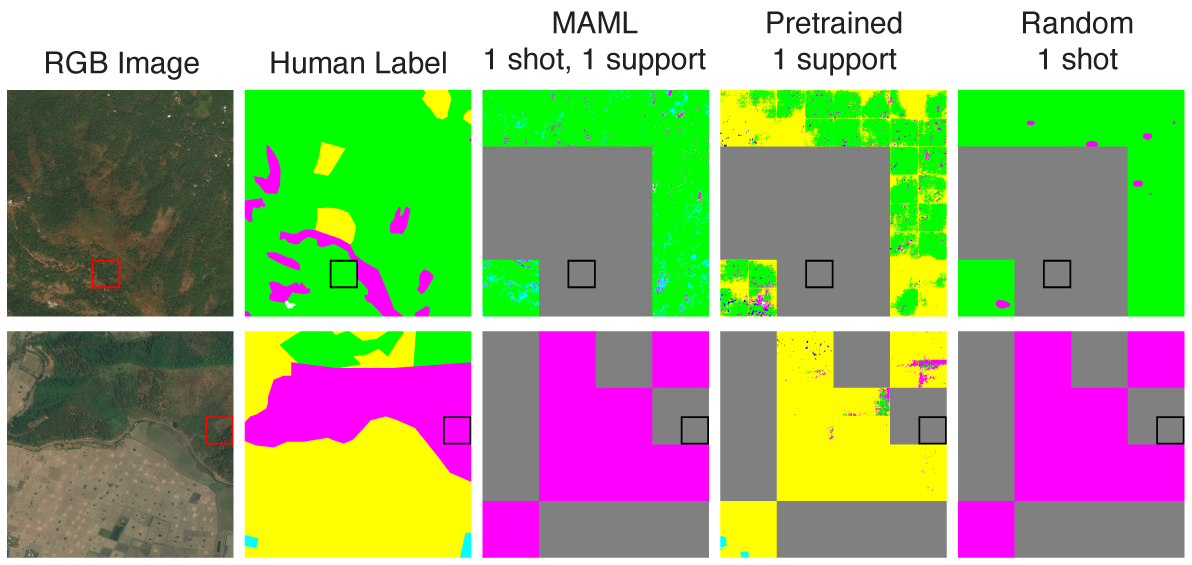}
		\caption{Clustered meta-dataset splits}
		\label{fig:deepglobe:examples_cluster}
	\end{subfigure}

	\ifdefined\reducecaptionspace
	\setlength{\belowcaptionskip}{-15pt}
	\fi

	\caption{Example segmentation predictions by MAML, a pre-trained U-Net, and U-Nets trained from scratch.}
	\label{fig:deepglobe_examples}	
\end{figure}
	
	\textbf{Random Meta-Dataset Split Results}.
	When the meta-datasets were randomly split, the pre-trained model performed better than MAML and the randomly initialized model. 
	This was especially true at smaller meta-training set sizes (\cref{fig:deepglobe:traintiles}).
	In other words, MAML requires a large set of meta-training tasks in order to perform well on new tasks. 
	As the number of shots seen by the meta-learner increases, MAML catches up to the pre-trained model (\cref{fig:deepglobe:shots}). 
	In these experiments, we did not observe fine-tuning of the pre-trained model to improve its performance.
	
	\Cref{fig:deepglobe:examples_random} visualizes the predictions of MAML and the baselines on 1-shot learning for two images: one where MAML performs well and one where it fails. 
	MAML appears heavily influenced by the choice of the 1 shot, while the pre-trained model is biased toward predicting agriculture (the most common class). 
	The model trained from scratch is even more heavily influenced by the choice of the 1 shot, as this is the only data it sees during training.
	
	The success of pre-training can be attributed to the complete overlap of meta-train and meta-test distributions, seen in \cref{fig:deepglobe:pca}. 
	In the setting where $p(X,y)$ are identical in the source domain and target domain, a model trained on the source domain transfers perfectly to the target domain.
	These results also expose MAML's weaknesses when meta-train size is small: it is not able to retain information about land cover types as effectively as a straightforward supervised model.
	
	\textbf{Clustered Meta-Dataset Split Results}. 
	When the meta-datasets were split along clusters, the meta-train and meta-test distributions overlapped less (\cref{fig:deepglobe:pca}) but could still be considered to arise from the same data-generating distribution. 
	Whereas the meta-train set contains mostly agriculture pixels, the meta-test set contains predominantly forest. 
	\Cref{fig:deepglobe:traintiles,fig:deepglobe:shots} show, first and foremost, that this meta-test set is more difficult than the randomly split meta-test set for all three models. 
	However, MAML is able to adapt to this distributional shift more successfully than the pre-trained model. 
	Example segmentations shown in \cref{fig:deepglobe:examples_cluster} reveal that MAML's flexibility to adaptation can again be both helpful and detrimental: helpful when the 1 shot is representative of the image, but detrimental when it is not.
	We see that the pre-trained model carries its bias toward agriculture into its meta-test set predictions, whereas MAML does not appear to retain a strong enough prior to recognize agriculture without being provided a shot containing that class.
	
	\subsection{Visualization of Model Adaptation}
	\label{sec:adaptation}

	\begin{figure}
		\begin{subfigure}[t]{\linewidth}
			\includegraphics[width=3.8cm]{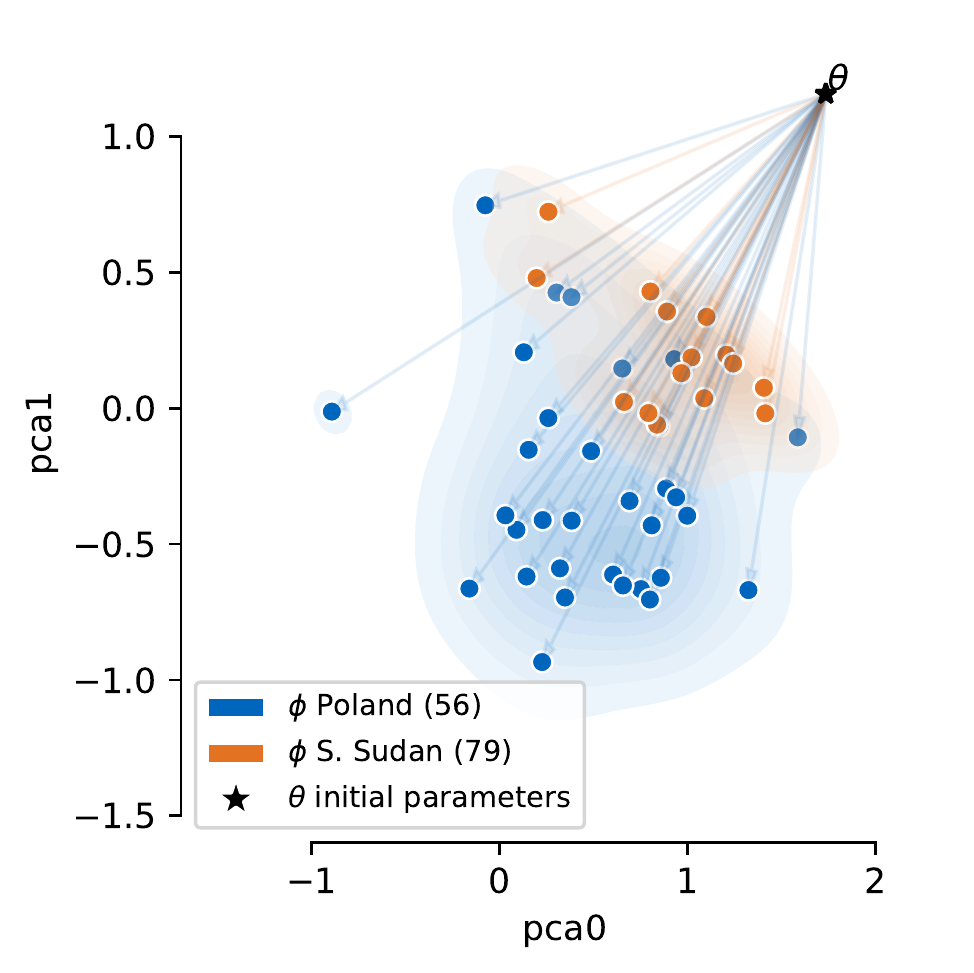}
			\includegraphics[width=3.8cm]{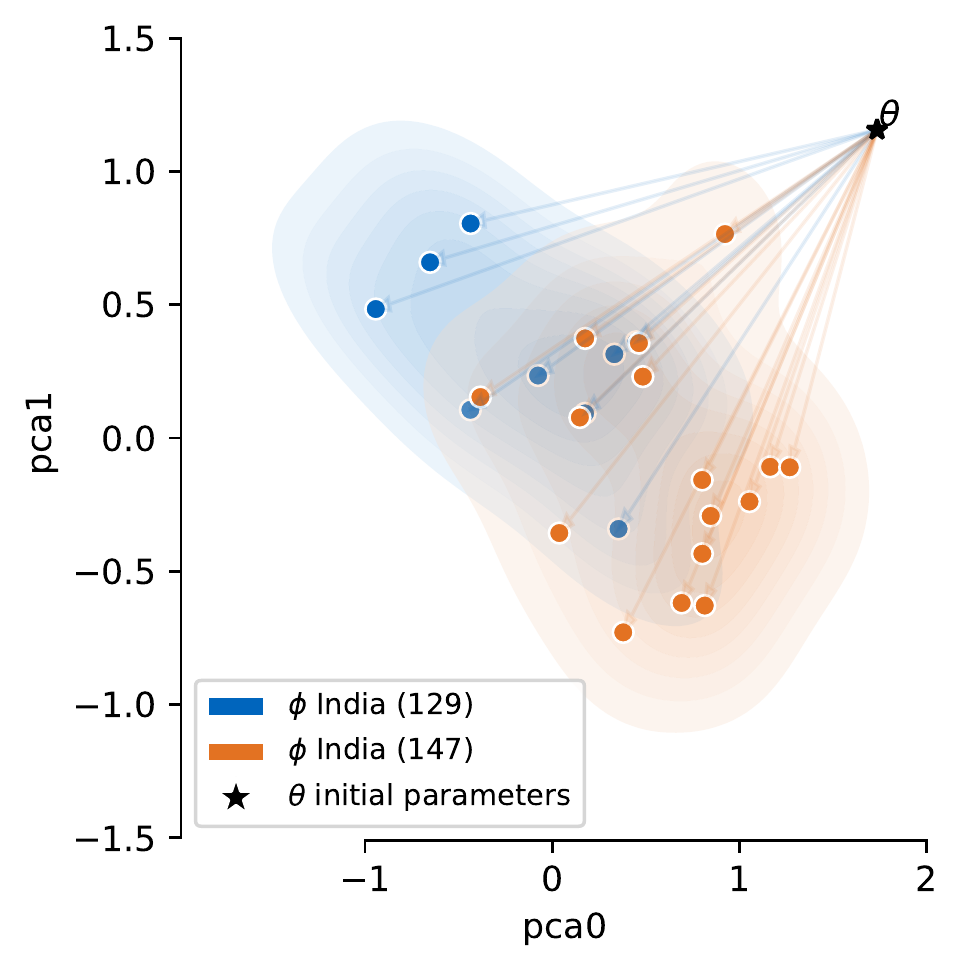}
			
			\includegraphics[width=3.8cm]{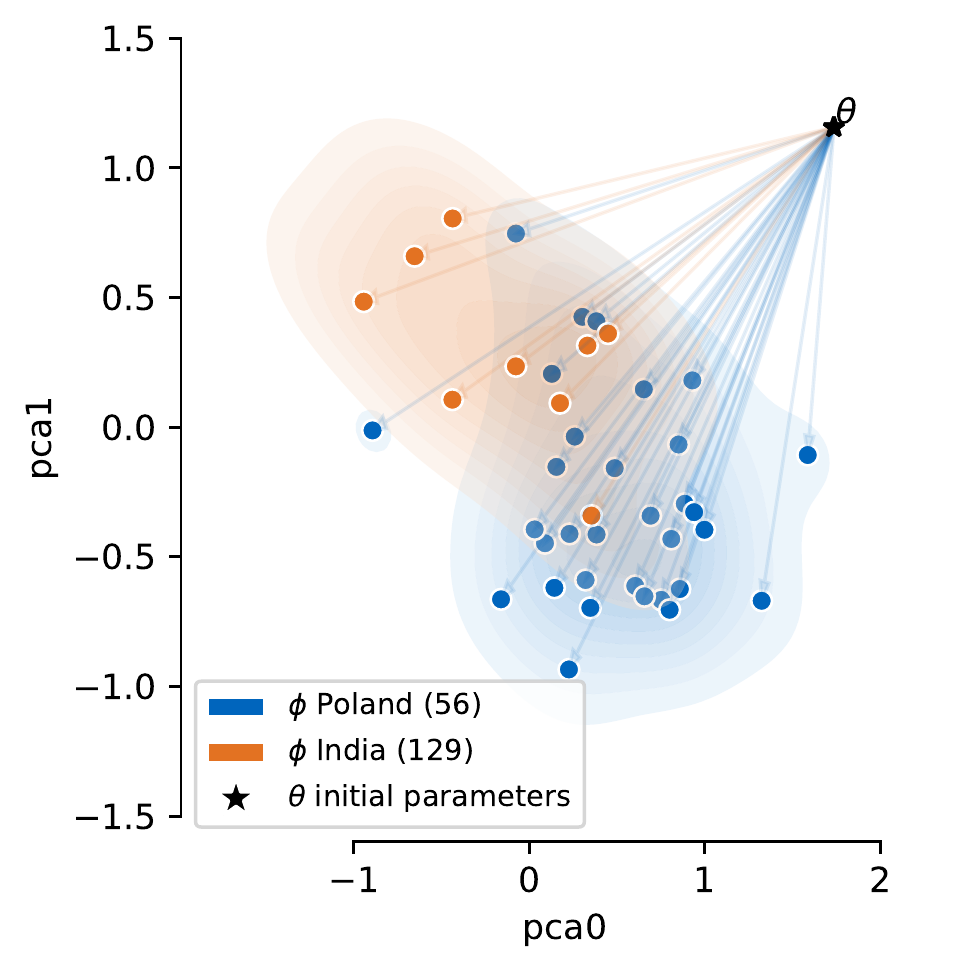}
			\includegraphics[width=3.8cm]{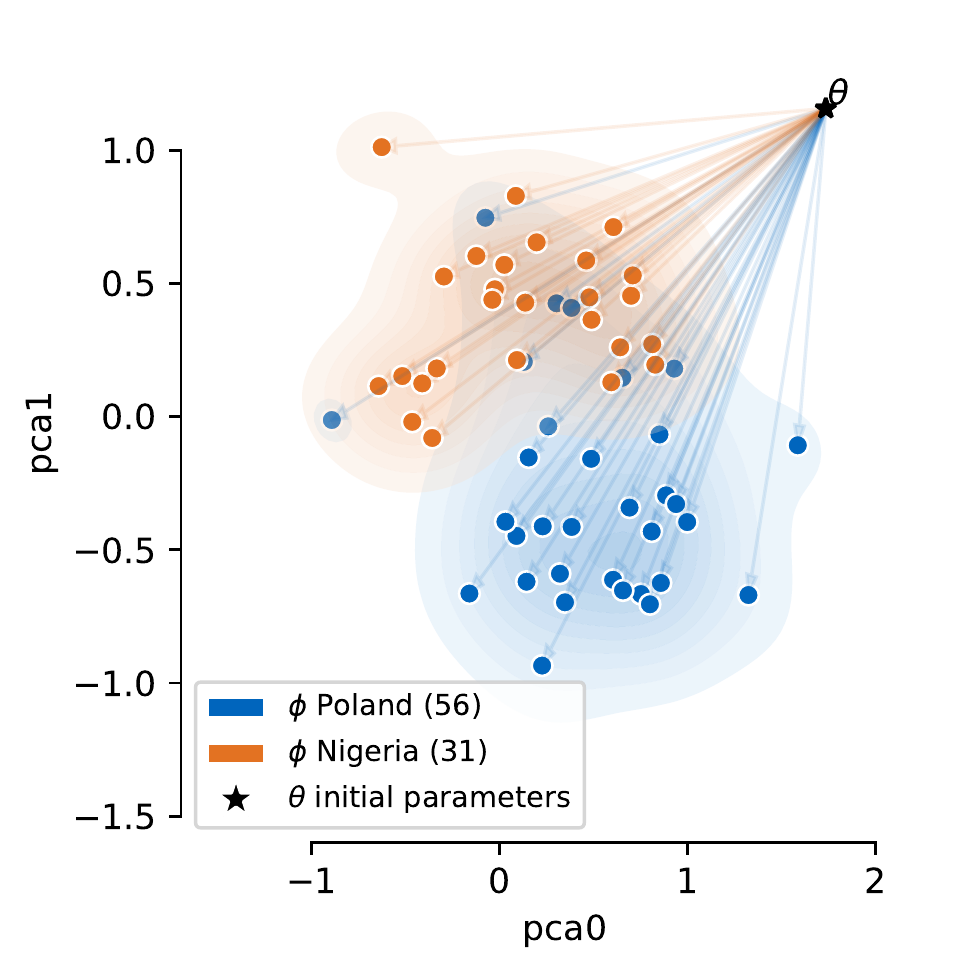}
			\caption{Adaptation of the MAML-trained CNN model to episodes from different regions.}
		\end{subfigure}

		\begin{subfigure}[t]{\linewidth}
			
			\includegraphics{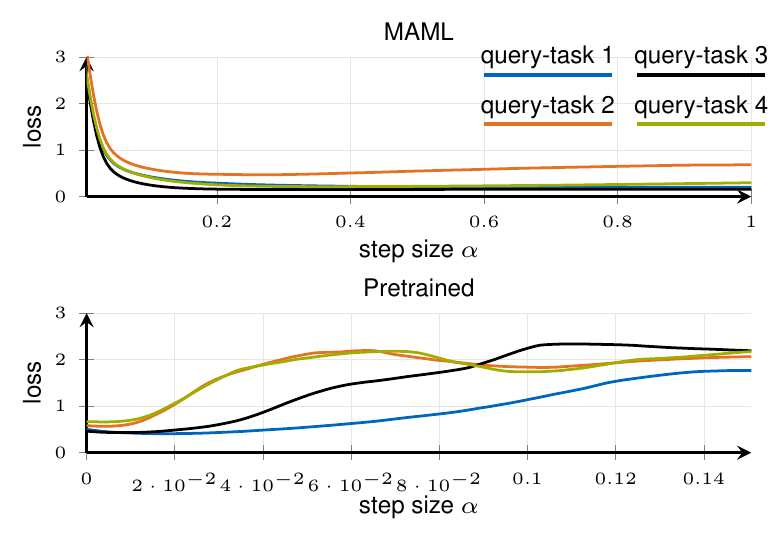}
			\tikzsetnextfilename{sen12ms-adaptation}

		\caption{1D Loss surface on multiple query-task samples along the gradient of one support-task}
		\label{fig:1dloss}
	\end{subfigure}
	
		
		
		
		

	\ifdefined\reducecaptionspace
	\setlength{\belowcaptionskip}{-15pt}
	\fi
	
	\caption{The adapted weights $\phi_\tau$ for task $\tau$ vary from region to region (a). The loss surface along the direction of initial weights $\theta$ to $\phi_\tau$ (b) is more convex and allows larger gradient step sizes for model-agnostic meta learning compared to regular pretraining.}
	\label{fig:adaptation}
\end{figure}

In the introduction and \cref{fig:representations}, we showed the regional diversity of representations on the Earth's surface using PCA on pre-trained VGG-16 image features.
In \cref{sec:metalearning} and \cref{fig:metalearning}, we assumed that a neural network would achieve optimal performance with a different set of weights $\phi^\ast$ for each geographic region.
In this experiment, we empirically confirmed this hypothesis with two evaluations on meta-test regions of the Sen12MS dataset.
In \cref{sec:adaptation:pca}, we visualize the adapted MAML weights for two distinct geographies.
Then in \cref{sec:adapataion:1dloss} we compare the loss surfaces of a MAML-trained and a pre-trained model along one adaptation trajectory. The two evaluations are meant to provide the reader with some intuition of what MAML is doing in different regions and how this differs from pre-training.

\subsubsection{Region-wise Adaptation}
\label{sec:adaptation:pca}
We studied the adaptation of MAML-model parameters $\theta$ trained on 2-shot 4-way tasks of \cref{sec:sen12msclassification}.
We sampled 1000 1-shot 4-way classification task-datasets from the meta-test regions for the four most common classes (forests, grassland, savanna, urban) and split these into a support and query partition at ratio of 4:1.
For each training task, we evaluated the gradient and adapted the model using gradient descent with step size $0.75$ to new parameters for each task $\phi^\tau$.
We visualized this adaptation by flattening all model parameters to a \num{231818}-dimensional vector and used PCA to map the parameters to the first two principal components.
We colored this embedding by region and drew lines from the initial weights $\theta$ to the adapted task-specific weights $\phi_\tau$ in \cref{fig:adaptation}.
The adapted model-weights differ characteristically between regions in embedding space, as can be seen in the examples of Poland and South Sudan.
This empirically shows that a separate set of model parameters is optimal for these two different regions.

\vspace{-.5em}
\subsubsection{Loss Surface along Support Gradient}
\label{sec:adapataion:1dloss}

Next, we evaluated the query loss along one line from initial parameters $\theta$ to task-adapted parameters $\phi$ with respect to four query tasks with the MAML-trained model and the regularly pre-trained model.
Here, we selected one support-task and four query-tasks from the same region and season.
The gradient $\V{g}$ was evaluated with respect to the support-task and obtained different model weights $\phi_\alpha$ along the gradient by $\phi_\alpha = \theta + \alpha\V{g}$ with different step sizes $\alpha_\text{MAML} \in [0,1]$, $\alpha_\text{pre} \in [0,0.15]$.
Initially, we evaluated larger stepsizes, but chose to show these intervals which are proportional to the optimal step sizes for MAML and pretrained model.
We calculated the query loss using the model $f_{\phi_{\alpha}}$ for each of the four test-tasks at different step sizes $\alpha$.
This draws a one-dimensional slice of the loss-surface along the gradient direction determined by the support-task.
In \cref{fig:1dloss}, we show this loss surface for the MAML-trained model and the pre-trained classification model. 
Without adaptation, at $\alpha=0$, the MAML-trained model evaluated a high loss compared to the pre-trained model.
This is consistent with the comparatively poor zero-shot results from \cref{fig:sen12ms-classification-segmentation}.
With increasing step size, however, we observe that the MAML loss decreased consistently while the pre-trained loss remained on similar level or increased for larger step sizes.
The MAML-trained model achieved low loss in a large range of step sizes from $0.1$ to $1$ through all query-tasks while a narrow range of step sizes between $0$ and $0.05$ lead to better accuracies on some tasks from the pre-trained model initialization.
In general, the loss surface of the MAML-trained model followed a convex curve for all of the test examples while the loss surface showed local minima from the pre-trained model initialization.
This experiment illustrates the difference between meta-learning and regular pre-training that lead to very different model parameters. 
The loss surface of a meta-learned model was more convex in the gradient direction of a novel task. 
Overall, the MAML-trained model benefited, \ie achieved lower test-loss, when being adapted to samples of a new region regardless of the step size. 
For the pre-trained model, it would have been beneficial not to adapt to one specific region for query tasks 2 and 4.

\section{Discussion and Conclusion}

In this work, we evaluated the model-agnostic meta-learning (MAML) algorithm for few-shot problems in land cover classification to adapt deep learning models to individual regions with few data examples.
Existing models use regular gradient descent to pre-train a model on a large body of data and use this pre-trained model as an initialization for datasets with fewer examples.
We compared these two approaches on land cover classification on the Sen12MS dataset of optical and radar images from globally distributed regions and the DeepGlobe dataset with very high-resolution imagery in few regions.
The results on Sen12MS in \cref{sec:sen12msclassification} demonstrate that MAML-optimization can outperform regular gradient descent and pre-training of models when the dataset includes a distinct regional diversity.
The DeepGlobe results in \cref{sec:deepglobe} illustrate the advantage MAML offers when the source domain differs from the target domain in transfer learning but also highlight MAML's weaknesses in retaining prior knowledge and under-performing in ideal (identical source and target domain) settings.
In \cref{sec:adaptation}, we evaluated the loss surfaces for pre-trained and MAML-trained models and showed that the loss surface was more convex for MAML-trained models when adapting to new unseen data.

We believe that the meta-learning framework can lead deep learning in Earth observation to a new direction: away from finding incrementally better model architectures for specific use-cases and toward unifying strategies that more closely reflect the reality on the Earth's surface. 
Much work remains to be done to improve MAML performance by retaining stronger priors on land cover classes, as well as to explore other meta-learning paradigms (\eg prototypical networks).

{\footnotesize
	\bibliographystyle{ieee_fullname}
	\bibliography{refs}
}

\end{document}